\pdfoutput=1
\documentclass[11pt]{article}
\usepackage{enumitem}

\usepackage{EMNLP2023}

\usepackage{pifont}

\usepackage{graphicx}

\usepackage{times}
\usepackage{latexsym}
\usepackage{booktabs}
\usepackage[T1]{fontenc}
\usepackage{tabularx}
\usepackage[utf8]{inputenc}

\usepackage{microtype}

\usepackage{inconsolata}

\newcommand{\OurDataset}{CommVQA}
\newcommand{\figref}[1]{Figure~\ref{#1}}
\newcommand{\secref}[1]{Section~\ref{#1}}

\usepackage{titlesec}
\titlespacing*{\paragraph}{0ex}{0.5ex}{1ex}

\title{\OurDataset: Situating Visual Question Answering\\in Communicative Contexts}

  \author{\normalsize Nandita Shankar Naik \\
  \normalsize Stanford University \\
  \normalsize \texttt{nanditan@cs.stanford.edu} \\\And
  \normalsize Christopher Potts \\
  \normalsize Stanford University \\
  \normalsize \texttt{cgpotts@stanford.edu} \\
\\\And
  \normalsize Elisa Kreiss \\
  \normalsize University of California, Los Angeles \\
  \normalsize \texttt{ekreiss@ucla.edu} \\}
  
\begin{document}
\maketitle
\begin{abstract}
Current visual question answering (VQA) models tend to be trained and evaluated on image-question pairs in isolation. However, the questions people ask are dependent on their informational needs and prior knowledge about the image content. To evaluate how situating images within naturalistic contexts shapes visual questions, we introduce \mbox{\OurDataset}, a VQA dataset consisting of images, image descriptions, real-world communicative scenarios where the image might appear (e.g., a travel website), and \textit{follow-up} questions and answers conditioned on the scenario and description. \OurDataset, which contains 1000 images and 8,949 question--answer pairs, poses a challenge for current models. Error analyses and a human-subjects study suggest that generated answers still contain high rates of hallucinations, fail to fittingly address unanswerable questions, and don't suitably reflect contextual information. 
Overall, we show that access to contextual information is essential for solving \OurDataset, leading to the highest performing VQA model and highlighting the relevance of situating systems within communicative scenarios.
\end{abstract}

\section{Introduction}

Visual question answering (VQA), the task of providing an answer to a question given an image, measures a model's ability to synthesize visual and textual modalities, and has many promising real-world applications. For example, when images online can't be seen and accessed, it severely affects people's abilities to educate themselves, socially engage, and stay informed \cite{morris2016most, macleod2017understanding, voykinska2016blind, gleason2019s}, and VQA models are an opportunity for providing interactive accessibility to such visual content at scale \cite{vizwiz, baker2021spoon}. While most prior VQA datasets focus on investigating image-text alignment as a decontextualized task \cite{vqa,vqa_v2,gqa, ok_vqa}, we aim to reframe it as a human-centric communicative problem, moving it closer to a real-world interactive setting.

According to a pragmatic Bayesian view of communicative actions, people tend to ask questions that maximize the contextually relevant information gain based on their existing prior beliefs about the world \cite{frank2012predicting}, following fundamental pragmatic principles \cite{grice1975logic}. Based on these linguistic insights, we argue that prior VQA datasets largely do not consider two central communicative drives that limit their utility.
The information people aim to obtain (and consequently the types of questions they ask) varies with (1) the person's \emph{information needs} based on their goals when encountering the image, and (2) the person's \emph{prior knowledge} of the image content. Thus we introduce \OurDataset, a benchmark that treats VQA as an inherently communicative task.\footnote{All data and code are available at: \url{https://github.com/nnaik39/commvqa}.}

\begin{figure*}
\includegraphics[width=.995\linewidth]{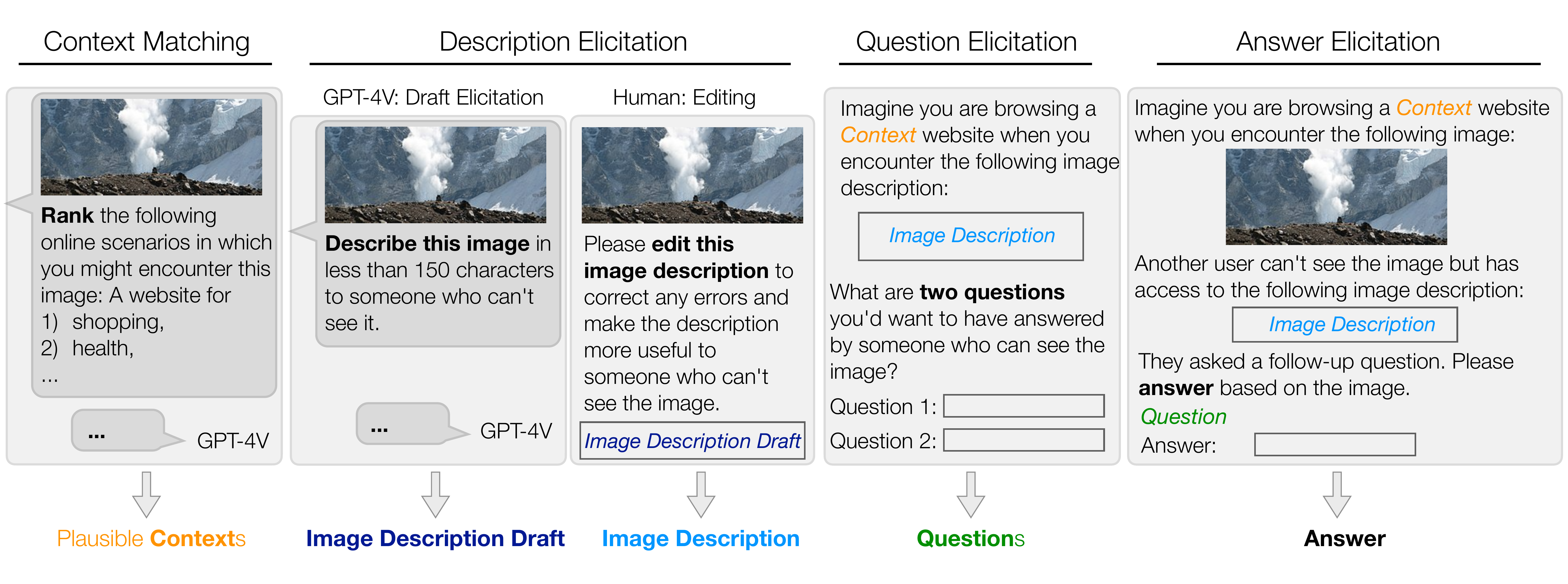}
  \caption{\textbf{Overview of the CommVQA Dataset Construction.} Images were sourced from Wikipedia and paired with relevant scenarios. The description were first generated by GPT-4V, then edited by humans. Other participants then provided questions and answers based on the scenario and description, resulting in at least three answers for each of the 2,983 unique visual questions. Simplified instructions are shown here; full details are in Appendix \ref{sec:appendix}.}
  \label{fig:dataset_overview}
\end{figure*}

To investigate people's \emph{information needs}, \OurDataset\ consists not only of images, questions, and answers, but also of image descriptions and plausible communicative scenarios for each image. Prior image accessibility research with blind and low-vision (BLV) participants shows that the information people want from an image is dependent upon the scenario in which the image appears \cite{beyondonesize, person_shoes_tree, kreiss2022context, muehlbradt2022s}. For example, if a person encounters an image when they are shopping online, they are likely to ask questions about the brands within the image, while if they're browsing the news, the perceived purpose of the image shifts, and they are likely to ask questions about the event occurring within the image \cite{beyondonesize, person_shoes_tree}. We therefore define a scenario as the type of website (e.g., a shopping website) combined with a goal for viewing it (e.g., to buy a gift) and expect it to shape people's information needs in a VQA task.

In addition, the relevant questions a person might ask are predicted to be guided by their \emph{prior knowledge} about the image. For most images we encounter online, we can rely on rich cues that allow sophisticated inferences about what an image contains. An image on a shopping website, for instance, would likely be accompanied by an article label, such as ``Colorful Summer Skirt'', or it could have an informative alt text description. In the standard VQA task, annotators are asked to write questions in isolation to ``fool a smart robot'', an adversarial task where the goal of the questioner is to trick an AI model \cite{vqa_v2}. However, in the naturalistic setting, these visual questions are better conceptualized as \emph{follow-up} questions, since they are conditioned on already available information. In \OurDataset, we situate the task by providing people with quality-controlled image descriptions instead of the image itself when collecting visual questions. Together with the contextual grounding of the images, this pipeline, presented in \figref{fig:dataset_overview}, generates a challenging dataset with context-sensitive, highly diverse questions, as well as longer answers compared to prior VQA datasets.

We benchmark four state-of-the-art VQA models on \OurDataset, and show that the situated nature of \OurDataset\ poses a significant challenge for current state-of-the-art vision-language models and allows for insights into model generation behavior due to a highly controlled data generation pipeline. We find that the most successful model needs to integrate contextual information, suggesting that context shapes VQA in communicative settings.

In summary, our main contributions are: (a) we introduce \OurDataset, a benchmark consisting of images, contexts, descriptions, questions, and answers, where models must infer the details most relevant to the questioner's goals, (b) we benchmark this dataset on current VQA approaches and explore whether models can be instructed to integrate context, and (c) we show via error analyses and a human-subjects experiment that even the best models generate false information at high rates.

\section{Related Work}

With \OurDataset, we aim to situate the abstract VQA task within communicative settings. This builds on prior VQA dataset work (\secref{sec:vqa}), and communicative insights from linguistics and human-computer interaction (\secref{sec:comm})

\subsection{VQA Datasets}\label{sec:vqa}

Most VQA datasets focus on image-question-answer triplets, which are constructed in isolation from the real-world contexts in which the images might appear \cite{vqa,vqa_v2,gqa, ok_vqa}. 
The VizWiz dataset \cite{vizwiz} stands out among VQA datasets as it focuses on real-world accessibility. The images in VizWiz are elicited from BLV users, who uploaded a picture to a phone app with a question about their real-world environment. The goals behind the questions and images is specific to BLV users exploring their physical environment. \mbox{\OurDataset} is complementary to VizWiz, since it investigates how contextually situating images online affects a model's VQA task performance. 

Some VQA datasets have explored integrating supplementary information. Visual Dialog \cite{visual_dialog} aims to create visual chatbots, which can answer a question based on an image and the prior dialogic context. Similarly to \OurDataset, these questions can be conceptualized as follow-up questions based on an image description. However, the images, questions, and answers are still decontextualized from a specific information goal beyond the questioner wanting to understand the image. \OurDataset\ instead extends the focus on varying the questioner's broader goals. 

ScienceQA \cite{lu2022learn} consists of multimodal science understanding questions, where each multiple choice question and answer is associated with a lecture and explanation, functioning as the context. VQAOnline \cite{chen2023fully} consists of images, questions, and context sourced from StackExchange, where the context is the body of the post. 
The notion of context in these datasets is limited to texts that provide supplementary information, whereas we specifically focus on the effect of changing scenarios on the VQA task. 

PromptCap \cite{hu2022promptcap} similarly considers the notion of ``context'' for models generating answers in the VQA task. First, the model generates a visual description relevant to the question and then generates an answer solely relying on the description, conceptualized as ``context.'' In contrast to PromptCap, the contextual condition in \OurDataset\ is strictly complementary to the image (see Section \ref{sec:imgrequired}), therefore fundamentally changing the task and modeling demands.

\subsection{VQA as Communication}\label{sec:comm}

Within pragmatics, there is a general consensus that questions are grounded in contexts and sensitive to the goals of the interlocutors \citep{Groenendijk84,Ginzburg96DYNAMICS,Roberts96,vanRooy03LP}.
In line with this prediction, prior human-subject studies with BLV participants show that context influences the information that people want about an image \cite{beyondonesize, person_shoes_tree, muehlbradt2022s, kreiss2022context}. 
\citet{beyondonesize} find that on social media, they wanted to know more about the people and the activity of the person who posted the image, while if people were on a shopping website to purchase a gift for a friend, they expressed a desire to learn more about the objects within the image. Inspired by \citet{beyondonesize}, we utilize a similar type of scenario. With \OurDataset, we constitute a large-scale dataset where context-sensitivity emerges with sighted participants in question and answer behavior and can be studied at scale.

\section{The CommVQA Dataset}

The \OurDataset\ dataset was constructed in five main steps, as visualized in \figref{fig:dataset_overview}. First, we sourced images from Wikipedia, and elicited both plausible scenarios and descriptions for the images from GPT-4V \cite{openai2024gpt4}. To ensure description quality, we conducted a human-subject study to edit the descriptions. We then elicited questions and answers for the dataset from US-based crowdworkers on Prolific. All human-subject studies were conducted with IRB approval.

\subsection{Image-Scenario Matching}

We extracted 1000 images from Wikipedia pages of topics related to at least one of the scenario conditions. We placed each image in two scenarios, which allows us to investigate how the scenario can induce variations in questions and answers across the same visual stimuli. Each image-scenario pair is annotated with on average 1.54 questions and each question with three answers, which results in 8,949 unique question-answer pairs. We prioritized high coverage of individual datapoints over breadth since the focus of this work is on uncovering the diverse contextual effects on the VQA task.

We chose six potential scenarios: Shopping, Travel, Science Magazines, News, Health, and Social Media. These scenarios were informed by prior work that showed people's information needs varied across these scenarios \cite{beyondonesize}, and allowed for an overlap between the images that appear in each scenario--for instance, images in travel blogs (e.g., a picture of a waterfall) could plausibly appear in an online science magazine.

To assign plausible scenarios to the image, we instructed GPT-4V \cite{openai2024gpt4} to rank them in order of descending plausibility, guided by the task in \citet{beyondonesize}.
We validated the assignment on a subset of data with a human-subject study before using GPT-4V to scale (see Appendix \ref{sec:tasks} for the prompt). From the top-three scenarios for each image, two scenarios were selected in order to balance out the co-occurrence of contexts. 

\subsection{Image Description Elicitation}
We elicited descriptions for all images in our dataset. They form the basis for the follow-up questions participants asked, simulating the effect of someone encountering an alt text associated with the image online. Importantly, descriptions were generated out-of-context so we could analyze context effects on the questions and answers without the description being a confounding factor.

\paragraph{Automatic Description Draft Elicitation}
We elicited initial description drafts by prompting GPT-4V with the phrase: \texttt{Describe this image in less than 150 characters to someone who cannot see it.} The length constraint follows a commonly-issued guide on best-practices for accessibility alt text writing \cite{alt_text_guide}. We chose Wikipedia for sourcing images due to the copyright permissions and image variety. All selected images were in the public domain.

\paragraph{Description Editing}

To ensure description quality, we conducted a human-subject study where participants edited the descriptions generated by GPT-4V \cite{openai2024gpt4}. These edits were intended to help balance for potential inaccuracies, possible misalignments with human description preferences, and the current design choices of GPT-4V. For instance, as of May 2024, the model refrains from explicitly identifying the people within an image \cite{gpt4_system_card}, even though proper names can be an important detail for a useful image description \cite{bennett2021s, macleod2017understanding}. 

Each participant was shown six randomized trials with an image and description. They were instructed to edit the image description to correct any errors and make it more useful to someone who cannot see it. Importantly, participants were not shown the context that each image was placed in to control for question variation across contexts. Participants could also choose to skip if no edits were needed. We recruited 369 participants and compensated them at the rate of \$13.50/hr.

We collected three description edits for each image description draft, and selected a random edited description for the final description.

\subsection{Question Elicitation}\label{sec:imgrequired}

To elicit visual questions, we recruited 619 participants, who were paid \$13.50/hr with an average completion time of seven minutes. In each trial, participants were given two pieces of information: an image description (e.g., ``A group of people of various ages walking along a grassy path, with trees on one side.''), and a scenario for the image (e.g., Imagine you are browsing a Health website, with the goal of learning how to live a healthier lifestyle). Crucially, participants didn't see the image to avoid priming for specific questions and simulate the visual inaccessibility of the image in the real-world scenario. They also rated how likely the image would appear within the provided scenario, and were prompted to ask two questions they would like answered by someone who can see the image.

In total, we elicited 2,983 questions, with 1.54 questions on average for each image-scenario pair. 
Based on a separate human-subject study (Section~\ref{sec:quantifying_hallucinations}), we find that 80\% of the questions require the image to answer, emphasizing the difficulty and inherent multimodal nature of \OurDataset.

\subsection{Answer Elicitation}

We elicited answers from 870 participants, who were paid \$13.50/hr, with an average completion time of seven and a half minutes. Participants were shown the image, question, context, and description. They were told that another user asked the question based on this image description, and asked to write an answer that would help the other person visualize the image. Each question was answered by at least three separate participants. 

We also included a checkbox for participants to indicate if a question was unanswerable. If a question was voted unanswerable by two or more annotators, we labeled the question as ``unanswerable'' within the dataset. In total, we collected 283 questions that were labeled unanswerable (9.5\%). Within the naturalistic setting, where people will ask questions about an image they cannot see, it's expected that people will ask unanswerable questions, and that models should have the capability to decline answering. The unanswerable portion of \OurDataset can help assess whether a model can abstain from answering instead of providing incorrect information \cite{macleod2017understanding}.

\section{Dataset Analysis}

The final \OurDataset\ dataset consists of 8,949 unique question-answer pairs, spanning across 2000 unique image-scenario pairs. To better understand the challenges posed by this dataset, we now provide an analysis of \mbox{\OurDataset}. Examples from the dataset, sampled to cover all scenarios, are given in Appendix Figure~\ref{fig:qualitative_examples}. We additionally conducted an analysis of a subset of the data and found similar patterns in model responses on this smaller dataset, suggesting our data is sufficiently large to obtain generalizable patterns. For more details, see Appendix \ref{sec:dataset_size_effects}.
 
\subsection{Analysis of Descriptions}

In \OurDataset, the descriptions are the basis for the VQA task and were collected in a two-stage process: first, automatic description generation by GPT-4V and then a human editing phase. The automatically generated descriptions had an average length of $63.663$ characters and in $42\%$ of trials, people didn't make any edits to the descriptions. When editing, participants added extra information, increasing the average length to $97.29$ characters.

\subsection{Analysis of Questions}\label{sec:questionanalysis}

\begin{figure}[tp]
\centering
\includegraphics[width=0.99\linewidth]{img/fig2.pdf}
  \caption{\textbf{Heatmap of BERT Classification Accuracy Across Scenario Pairs.} When fine-tuned on different scenario pairs, BERT exhibits varying performance in its ability to distinguish between these scenarios. For instance, BERT achieved 94\% accuracy when distinguishing between science magazines and shopping, but only $83\%$ accuracy for travel and social media.}
  \label{fig:heatmap}
\end{figure}

The main goal of \OurDataset\ is to situate the VQA task in a communicative context, assuming that the context shapes what becomes relevant and therefore has implications for downstream model performance. In this section, we investigate the context-sensitivity of the questions in \OurDataset.

If questions are context-sensitive, then a trained classifier should be able to predict the correct scenario from the question and achieve a performance reliably higher than random. To investigate whether this is true for \OurDataset, we fine-tuned BERT \cite{devlin2018bert} on the task of predicting whether each question appears within a certain context or not. We expect this task to be difficult even for a human, given that questions such as ``What time of day is it?'' might appear in any scenario. We split all questions in the dataset into an 80-10-10 train/test/val split, and fine-tuned for 200 epochs with LoRA \cite{hu2021lora}. 
Fine-tuned BERT achieved an accuracy of $56\%$ on this task, a significant improvement over random choice (16\%), showing that questions elicited within different contexts are inherently distinct.

While we find overall evidence that questions vary between all scenarios, a classifier analysis further allows us to investigate which individual scenarios have the least and most overlap in the questions asked. Figure \ref{fig:heatmap} shows the BERT classifier's performance when fine-tuned on distinguishing pairs of scenarios (e.g., shopping from science magazines), and indicates which scenarios are more easily distinguishable.
Intuitively, certain scenarios are more related than others--for instance, shopping and social media content is more likely to contain images of famous models, while shopping and science magazines are less likely to have overlap. Our analysis confirms this intuition and highlights where models may fail to translate across scenarios.

We further inspect the question interrogatives for a potential source of between-scenario variation. The results of a Welch's t-test demonstrated that, for example, compared to all other scenarios, ``Who'' questions are significantly more likely to appear in the social media scenario ($t(589.6) = 3.85$, $p < 0.001$), and ``Where'' questions are significantly more likely to appear in the travel scenario ($t(627.3) = 3.58$, $p < 0.0001$) compared to any other scenario. These results indicate that different question types are more likely to be asked in certain scenarios, which carries implications that model evaluation should be contextual. A model that has poor performance on ``Who'' questions might seem competent in non-social media scenarios, but fail to generalize to social media due to the distinct nature of the user's information needs. However, question type and scenario are still distinct conceptually. 

Taken together, we find converging evidence that the questions in \OurDataset\ fundamentally vary based on the scenario the images were presented in, highlighting the diverse requirements for building robust communicatively situated VQA models.

\subsection{Analysis of Answers}

We now turn to a general analysis of the \emph{answers} in \OurDataset\ and investigate the extent to which they are contextually situated.

Firstly, with an average length of $10.98$ words, the answers in \OurDataset\ are surprisingly long compared to prior datasets that used a similar online answer elicitation setup (e.g., 1.1 words for VQA-v2 \cite{vqa_v2} and 2.1 words for Visual Dialog \cite{visual_dialog}). Contrasting prior work, the instructions of \OurDataset\ are framed as a communicative task, where answerers were asked to help someone else who cannot see the image. It's plausible that the longer responses are partially due to the participant's wish to faithfully communicate with the questioner \cite{grice1975logic}.

\begin{figure}
\includegraphics[width=220px]{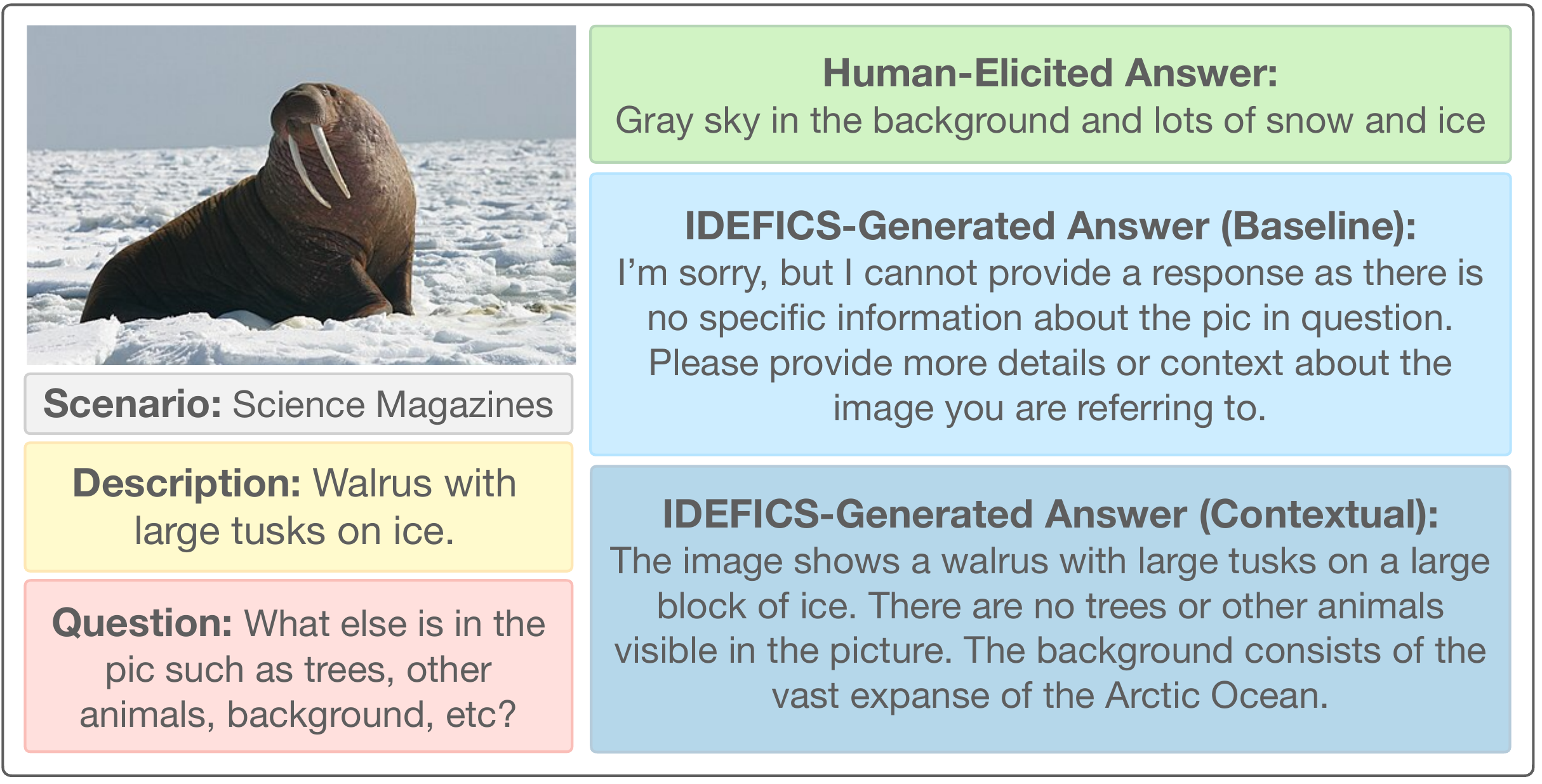}
\caption{\textbf{Example of Context Dependency in Answer Generation.} In this example, the question explicitly asks for content that is not in the description. While the human-elicited answers do not repeat information in the description, IDEFICS (contextual) provides an answer, but does repeat content that is in the description.}
\label{fig:answer_variance_example}
\end{figure}

While we focus our analysis on the effects context directly has on the questions being asked (see Section~\ref{sec:questionanalysis}), we also find that access to context becomes strictly necessary to answer those contextualized questions. Most strikingly, questions of the form ``What else is in the image?'' directly require the answerer to know what information users already have (see Figure~\ref{fig:answer_variance_example} and Figure~\ref{fig:multiple_answer_variance_examples} for examples).

\section{Model Benchmarking}

\begin{table*}
\centering
\begin{tabular}{llllll}
        \toprule
Model         & BLEU-1 & BLEU-4 & METEOR & ROUGE & CIDEr \\
\cmidrule{1-6}
IDEFICS                & 0.273 & 0.084 & 0.179 & 0.378 & 0.758 \\ %& 0.679 \\
IDEFICS (Contextual)   & \textbf{0.285} & \textbf{0.092} & \textbf{0.195} & \textbf{0.392} & \textbf{0.839} \\ % & 0.691 \\
\cmidrule{1-6}
LLaVA              & 0.195 & 0.052 & 0.172 & 0.317 & 0.453 \\%& 0.731 \\
LLaVA (Contextual) & 0.152 & 0.037 & 0.161 & 0.253 & 0.219 \\%& 0.786 \\ 
\cmidrule{1-6}
mPLUG-Owl              & 0.199 & 0.048 & 0.169 & 0.320 & 0.462 \\%& 0.720 \\
mPLUG-Owl (Contextual) & 0.187 & 0.046 & 0.172 & 0.295  & 0.355 \\%& 0.758\\
\cmidrule{1-6}
BLIP-2                 & 0.267 & 0.059 & 0.098 & 0.282 & 0.434 \\%& 0.613\\
BLIP-2 (Contextual)    & 0.015 & 0.001 & 0.028 & 0.043 & 0.014 \\%& 0.586\\
        \bottomrule
\end{tabular}
\caption{\textbf{Comparison of Baseline and Contextual Conditions Across Models.} This table presents results for both baseline and contextual conditions across all models. IDEFICS (contextual) achieved the highest scores across all metrics. Results are averaged over three random data splits and model-generated answers.}
\label{fig:results_table}
\end{table*}

In this section, we investigate the performance of four state-of-the-art vision-language models on \OurDataset\, and to what extent providing context to the models improves their performance.

We selected four models to benchmark on this dataset: LLaVA \cite{liu2024visual}, BLIP-2 \cite{blip2}, mPLUG-OWL \cite{ye2023mplugowl}, and IDEFICS \cite{laurençon2023obelics}. To maximally enable the reproducibility of this work and support the public development of models, we focus on open-source models. LLaVA is trained on a multimodal instruction-following dataset, and exhibits strong performance across several VQA datasets \cite{vqa_v2, gqa}. mPLUG-Owl \cite{ye2023mplugowl} is also instruction-tuned, and displays competitive performance on various VQA benchmarks, including VQA-v2 \cite{vqa_v2}. IDEFICS \cite{laurençon2023obelics} is an open-access reproduction of Flamingo \cite{alayrac2022flamingo}, and was trained on a naturalistic web-scale dataset of interleaved image-text documents, including 141 million web pages. We selected LLaVA, mPLUG-Owl, and IDEFICS since their instruction-tuned nature allows us to straightforwardly integrate context. We also included a non-instruction-tuned model, BLIP-2, a general-purpose vision-language model that is additionally fine-tuned on the VQA task.

\paragraph{Evaluation Metrics}
We used three types of evaluation methods for complementary insights on the quality of the model-generated answers. To obtain insights on the overall string similarity of the model-generated and the human-written answers, we use the well-established reference-based NLG metrics BLEU \cite{bleu}, CIDEr \cite{cider}, METEOR \cite{meteor}, and ROUGE \cite{lin2004rouge}. However, these metrics do not adequately capture object hallucinations \cite{objectHallucination}, which is why we supplement this analysis with a human-subject study specifically aimed at capturing hallucinated content.
Finally, to measure how closely an answer's visual details matches with the image and associated visual description, we also include an analysis using CLIPScore \cite{hessel2021clipscore}.

\paragraph{Experimental Setup}

For all models, we conducted two main experiments. The \emph{contextual} condition simulated the situation human participants were in, where the models had complete access to the image, description and situational scenario while answering the question.

We include our prompt for the contextual condition in Appendix \ref{sec:tasks}. In addition, we tested a \emph{baseline} condition where the models only had access to the image and question, in order to assess the model capabilities for incorporating contextual information. All evaluations were conducted with the greedy decoding method to ensure reproducibility.

\paragraph{Overall Results}

In Table \ref{fig:results_table}, we show the performance of all models when provided with the image and question (baseline) and the full context that was available to human answerers (contextual). 
The IDEFICS model with the contextual condition has the highest performance for all models across all metrics. Evidently, integrating contextual information made the content itself more aligned with the ground-truth answer references, as evaluated by the reference-based metrics. 
While IDEFICS 
significantly improves when prompted with the contextual condition, this pattern is reversed for all other models. We hypothesize that less well-performing models closely reiterate the visual information they receive from the image description rather than adding new information to answer the question, which we turn to next.

\subsection{Repetition of Visual Information Within Generated Answers}

In order to investigate how much visual information models re-iterate within their generated answers, we conducted a similarity analysis of the model-generated answers to the image based on CLIPScore \cite{hessel2021clipscore} and a similarity analysis of the generated answers to the human-written image descriptions using Sentence-BERT embeddings \cite{sbert}. The analyses show convergent results.

While the reference-based metrics tend to decrease in the contextual condition, CLIPScore ratings (i.e., the similarity of the generated answer to the image) largely increase, as seen in Table \ref{fig:clipscore_results_table}. The only exception is BLIP-2, which significantly deteriorates in performance across metrics when contextual information is provided, largely rendering it uninterpretable.

Similarly, Figure \ref{fig:cosine_sim_plot} shows that the similarity between the description and the generated answers significantly increases with the contextual condition across models, according to a two-sample t-test analysis.
However, the increase in description similarity between the baseline and contextual condition is the lowest for IDEFICS. We conclude that in their answers, the contextual versions of LLaVA, mPLUG-Owl, and BLIP-2 mimic the description more than IDEFICS does, which explains the CLIPScore increase across the baseline and contextual conditions for these three models. These findings suggest that those models might overly emphasize the descriptive details when available, leading to less alignment with the ground-truth answers.

\subsection{Quantifying Hallucinations}
\label{sec:quantifying_hallucinations}

While IDEFICS (contextual) best approximates the human answers, it's still far from achieving human-like performance. When a model is intended to make information available that can't be verified by a user, it is especially important that this model only generates truthful content \cite{macleod2017understanding}. While this is straightforward to evaluate for accuracy-based VQA datasets, this isn't easily captured by the similarity-based metrics that are used for long-form evaluation, as measured by BLEU and CIDEr. In this section, we aim to give a brief intuition about the rate of hallucinated or incorrect pieces of information contained in the answers of the best-performing model, IDEFICS (contextual).

\begin{table}
\centering
\begin{tabular}{lcc}
        \toprule
Model         & CLIPScore & CLIPScore (C.) \\
\midrule
IDEFICS                & 0.679 & 0.691 \\
LLaVA              & \textbf{0.731} & \textbf{0.786} \\
mPLUG-Owl              & 0.720 & 0.758 \\
BLIP-2                 & 0.613 & 0.586\\
        \bottomrule
\end{tabular}
\caption{\textbf{CLIPScore Improvement Across Baseline and Contextual Conditions.} When each model is prompted with the contextual condition, CLIPScore increases, indicating that the model is repeating visual details. ``C.''\ represents the contextual condition.}
\label{fig:clipscore_results_table}
\end{table}

To estimate how many answers contain wrong, hallucinated or unverifiable content generated by the model, we conducted a human-subject study where participants were asked to rate model-generated answers. We randomly selected 100 answers from IDEFICS (contextual) to provide an assessment of the best-performing model. 70 participants were recruited, and compensated at a rate of \$13.80/hr. For each image, participants were asked to rate whether each answer contained information that was clearly not within the image, and to evaluate whether the image was strictly necessary for providing an answer to the question.

Overall, participants indicated that 23\% of the model-generated answers contained clearly erroneous information (with a Fleiss kappa inter-annotator agreement of 0.47) and for another 22\%, the truthfulness of the answers couldn't be clearly established. These results indicate that even the best-performing model (IDEFICS (contextual)) generates a high degree of erroneous information, making it unreliable for downstream use.

\subsection{Evaluating Unanswerability}

Since the questions in \OurDataset\ are asked by people who cannot see the image, these questions are not always answerable. When a person asks an unanswerable question, ideally, models should abstain from answering rather than provide incorrect, hallucinated information \cite{whitehead2022reliable}. 

We evaluated all models on the 238 unanswerable questions in \OurDataset. For each question, we assessed whether the model was able to successfully abstain from answering or if the model provided a hallucinated answer. To classify responses as ``Abstention'' or ``Non-Abstention,'' we conducted a direct string matching analysis to search for language that models commonly use to abstain (e.g., ``I cannot answer.'') A full list of strings in this analysis is provided in Appendix \ref{sec:unanswerability_strings}.

\begin{figure}
\centering
\includegraphics[width=0.99\linewidth]{img/fig4.pdf}
\caption{\textbf{SBert Cosine Similarity Analysis in Human and Model Responses.} 
Significance levels are marked with asterisks based on a two-sample t-test analysis.}
\label{fig:cosine_sim_plot}
\end{figure}

Overall, IDEFICS (contextual) had the highest rate of successful abstentions at 21\%, and BLIP-2 had the lowest rate, at 0\% (in both contextual and baseline conditions). Without access to contextual information, IDEFICS's abstention rate drops to 14\%, suggesting that context is helpful for accurately identifying unanswerable questions. Appending the string ``If you don't know, say `unanswerable''' to the model prompt improved the rate of IDEFICS' (contextual) abstention performances on the unanswerable questions to 87\%, suggesting that implicit instructions could help models abstain when necessary.

Questions most human participants judged as \emph{answerable} tended to be answered by the models as well. Without the contextual condition, LLaVA and IDEFICS declined to answer for answerable questions at a rate of 8.3\% and 8.6\%, respectively. For both models, the rate of false positives decreased in the contextual condition (LLaVa: 7.6\%; IDEFICS: 7.1\%).

Taken together, these results highlight that \OurDataset\ poses a challenging problem for state-of-the-art models. Our analyses suggest that models might not be able to leverage contextual information effectively and the high degree of hallucinations makes them unreliable, highlighting two important areas for future research.

\section{Conclusion}

Visual question answering models are a promising tool for making visual content accessible to all. With \OurDataset, we move the contextually isolated VQA task into a communicative setting that starts reflecting the diversity of downstream use, while keeping close control over the nature of the dataset to ease interpretability. We find strong evidence from dataset analysis that the types of relevant questions and answers change with the contextual domains where images appear. We also find evidence that integrating contextually relevant information improves model performance. Our results suggest that the path towards building viable VQA systems requires a focus on the wider communicative context where images appear.

\section*{Limitations}

In this work, we show that the scenarios images are presented in fundamentally affect the VQA task. To investigate this, we varied the broad type of website where we embedded the image (Social Media, Shopping, etc). However, context effects are likely much more diverse than the effects studied in this work. For example, recent work suggests that even topic changes within a website domain (a Wikipedia article on Mountains vs.~Body of Water) change the information needs that sighted and BLV users have for image descriptions \cite{refless_metrics}. This result likely translates to the VQA task and needs further investigation.

To allow for easy manipulations of the context domains and a highly controlled recruitment, participants were put in simulated scenarios where they were told about the website domain where the image appears. While even in these induced contextual setups, we find significant contextual variations, future work needs to explore the way this extends to real-world user experience outside of simulated scenarios.

In \OurDataset, we elicited questions and answers from sighted participants. Evidence from prior work \cite{beyondonesize} indicates that these results would likely transfer more directly to the accessibility scenario, but future work needs to analyze how the sighted user behavior translates to the BLV population more directly.

During the image description generation phase, initial descriptions are generated by GPT-4 and then edited by humans. This raises the potential issue that humans may be biased towards the GPT-4 responses, including keeping in details that may not be factually accurate. There is in fact evidence that starting out from an alt text shapes what type of descriptions people write and the quality of them \cite{mack2021designing}. Work on alt text captioning has primarily focused on old systems, which were far less competent, and more research is now needed to investigate how that interacts with newer systems. In our dataset, the descriptions themselves are not proposed as potentially ideal descriptions for accessibility. They simply provide the contextual framing for the questions after. 
% Entries for the entire Anthology, followed by custom entries
\bibliography{anthology,custom}
\bibliographystyle{acl_natbib}

\appendix
\newpage 

\section{Appendix}
\label{sec:appendix}

\subsection{Statistical Comparisons with Other VQA Datasets}

In Table \ref{table:statistical_comparison}, we provide a statistical comparison of CommVQA with other VQA datasets.

\begin{table*}
\centering
\begin{tabular}{p{2cm}p{2.5cm}p{1.5cm}p{1.5cm}p{1.5cm}p{1cm}p{1cm}p{1cm}}
        \toprule
Dataset         & Who Asked? & $\overline{Q}$ & $\overline{A}$ & $\overline{C}$ & \# Imgs & \# Scenarios & \# QA Pairs \\
\cmidrule{1-8}
VQA-v2 & Crowdworkers & 6.1 & 1.2 & \ding{55} & 204,700 & 1 & 658,111 \\
VizWiz-VQA  & Blind people &  6.7 & 1.7 & \ding{55} & 20,500 & 1 & 31,000 \\
OK-VQA    & Crowdworkers & 8.1 & 1.3 & \ding{55} & 14,031  & 1 & 14,055 \\
DocVQA    & Remote workers & 9.5 & 2.4 & \ding{55} & & 1 & \\
\cmidrule{1-8}
Visual Dialog$^*$              & Crowdworkers & 5.1 & 2.9 & \ding{55} &120,000 & 1 & 1.2M \\
VQAOnline  & Stack Overflow users & 9.3 & 173.2 & 127 & 64,700	& 1 & 64,700 \\
ScienceQA  & Elementary and high school curricula & 12.1 & 4.4 & 41 & 6,500 & 1 & 21,208 \\
CommVQA (ours)    & Crowdworkers who can only see the description & 7.3 & 10.98 & 17.2 & 1000 & 6 & 8949 \\
        \bottomrule
\end{tabular}
\caption{\textbf{Statistical Comparison: CommVQA and Other VQA Datasets.} This table compares CommVQA with seven other VQA datasets. The first four rows cover image-question-answer inputs, while the bottom three rows cover image-question-answer-context inputs. $\overline{Q}$,  $\overline{A}$, and $\overline{C}$ denote the average answer, question, and context lengths (as measured by the number of words), respectively. *Since Visual Dialog contains multi-turn conversations, we only included the average question length for the first round in the conversation.}
\label{table:statistical_comparison}
\end{table*}

\subsection{Dataset Collection Overview}
\label{sec:tasks}
In this section, we provide a full list of the tasks and prompts provided in each step of the dataset collection process.

For the scenario matching stage, we prompted GPT-4V with this prompt: \\
\texttt{Imagine you are a person browsing the Internet. Please rank the following scenarios in which you might encounter this image: \\ 1) You are browsing a shopping website, with the goal of purchasing an item or experience. \\
    2) You are browsing science magazines (such as National Geographic), with the goal of learning more about recent science developments. \\
    3) You are browsing news websites (such as New York Times), with the goal of learning more about recent news developments. \\
    4) You are browsing a health website, with the goal of learning how to live a healthier lifestyle. \\
    5) You are browsing social media, with the goal of learning more about your connections. \\
    6) You are browsing a travel website, with the goal of traveling to a new location.}

For the description elicitation, we asked participants to: ``Please edit this description of the image to correct any errors and make the description more useful to someone who cannot see it.'' This phrase was intended to incentivize people to both fix errors and include any communicative details that they felt was missing.

For the question elicitation study, we asked participants: ``Imagine you are browsing a \textbf{scenario} website when you encounter the following image: \textbf{description}. If you encounter this image on a \textbf{scenario} website, what are two questions you'd want to have answered by someone who can see the image?''

For the answer elicitation study, we asked participants to: ``Imagine you are browsing a \textbf{scenario} website when you encounter the following image. \textbf{image} Another user cannot see the image directly but has access to the following image description: \textbf{description}. Based on the description, they asked a follow-up question. Please answer based on the image: \textbf{question}.''

We recruited annotators through Prolific [3], an online annotation platform similar to Amazon’s Mechanical Turk. We recruited only US-based participants, and the only prescreener we used was to exclude participants who had taken a previous stage of the description generation process. For instance, participants who edited descriptions were excluded from providing questions or answers.

\subsection{Context Integration Prompts for Models}

We integrated context by prompting models with the format:

\texttt{Assume someone is browsing a \{scenario\} website when they encounter this image. They cannot see the image directly, but they can access this image description: \{description\}. Based on this description, they asked this follow-up question. Please answer based on the image. In your answer, prioritize details relevant to this person. Question: \{question\}}

\subsection{Additional Dataset Composition Analysis}

\begin{figure*}
\includegraphics[width=.99\linewidth]{img/fig5.pdf}
\caption{\textbf{CommVQA Dataset Examples.} Four example entries from the CommVQA dataset, each paired with a randomly selected answer.}
\label{fig:qualitative_examples}
\end{figure*}

\begin{figure*}
\includegraphics[width=.99\linewidth]{img/fig6.pdf}
\caption{\textbf{Examples of Model-Generated Answers.} This figure shows examples of dataset entries and model responses from IDEFICS, the leading model. Sample responses from IDEFICS, the leading model, indicate that context aids in producing answers that better match human ground-truths, despite some limitations in context integration.}
\label{fig:multiple_answer_variance_examples}
\end{figure*}

\subsection{Models Chosen}
We evaluated the following model versions, which are freely available on HuggingFace:

\paragraph{LLaVA} llava-v1.5-14b-3GB
\paragraph{IDEFICS} idefics-9b-instruct
\paragraph{mPLUG-Owl} mPLUG-owl-llama-7b
\paragraph{BLIP-2} blip2-opt-2.7b

\section{Unanswerability Analysis}
\label{sec:unanswerability_strings}

In this section, we include a list of the strings used to evaluate whether a model abstained from answering.

\begin{itemize}[noitemsep]
    \item \texttt{I cannot answer}
    \item \texttt{I cannot determine}
    \item \texttt{I cannot provide}
    \item \texttt{unanswerable}
    \item \texttt{I don't have enough information}
    \item \texttt{AI language model}
    \item \texttt{I don't have the context}
    \item \texttt{I don't have enough context}
    \item \texttt{I'm sorry}
    \item \texttt{I don't have the capability}
    \item \texttt{I don't have enough information}
\end{itemize}

\begin{table*}
\centering
\begin{tabular}{llllll}
        \toprule
Model         & BLEU-1 & BLEU-4 & METEOR & ROUGE & CIDEr \\
\cmidrule{1-6}
IDEFICS       & 0.326 & 0.106 & 0.189 & 0.401 & 0.825 \\
IDEFICS (Contextual)   & \textbf{0.356} & \textbf{0.122} & \textbf{0.198} & \textbf{0.413} & \textbf{0.904}\\
\cmidrule{1-6}
mPLUG-Owl     & 0.191 & 0.045 & 0.166 & 0.311 & 0.414  \\
mPLUG-Owl (Contextual) & 0.178 & 0.043 & 0.169 & 0.284 & 0.335  \\
\cmidrule{1-6}
LLaVA         &  0.192 & 0.050 & 0.171 & 0.311 & 0.431 \\
LLaVA (Contextual)     & 0.149 & 0.037 & 0.161 & 0.247 & 0.200  \\ 
\cmidrule{1-6}
BLIP-2        & 0.289 & 0.067 & 0.100 & 0.286 & 0.435 \\
BLIP-2 (Contextual)    & 0.059 & 0.010 & 0.078 & 0.224 & 0.343 \\
        \bottomrule
\end{tabular}
  \caption{\textbf{Model Benchmarking Results on a Subset of 150 Images from CommVQA.} Reflecting similar results from the full dataset benchmark, the highest-scoring model is IDEFICS (contextual), and LLaVA and mPLUG-Owl both exhibit decreased performance within the contextual condition.}
\label{table:subset_results_table}
\end{table*}

\section{Effects of Dataset Size}
\label{sec:dataset_size_effects}

We conducted an analysis on a subset of our dataset and uncovered similar effects of model behavior. In Table \ref{table:subset_results_table}, we present model benchmarking results on a subset of 150 images and 1,215 unique visual questions within our dataset. Largely, we surface similar patterns as found within the full dataset. Namely, IDEFICS (contextual) exhibits the highest performance, and LLaVA and mPLUG-Owl both decrease their performance with their contextual condition, but this is explained by an increase in CLIPScore, as shown in Table \ref{table:subset_results_clipscore}. 

\begin{table}
\centering
\begin{tabular}{lll}
        \toprule
Model         & CLIPScore & CLIPScore (C.) \\
\cmidrule{1-3}
\cmidrule{1-3}
IDEFICS                & 0.667 & 0.666 \\
\cmidrule{1-3}
LLaVA              & \textbf{0.729} & \textbf{0.784} \\
\cmidrule{1-3}
mPLUG-Owl              & 0.720 & 0.756 \\
\cmidrule{1-3}
BLIP-2                 & 0.604 & 0.662\\
        \bottomrule
\end{tabular}
\caption{\textbf{CLIPScore Improvement Across Baseline and Contextual Conditions for Subset of 150 images.} This table displays the increase in CLIPScore for each model when comparing baseline and contextual conditions for the subset of 150 images, finding a similar trend as when this result is run on the full dataset. In particular, LLaVA (C.) has the highest CLIPScore condition, and the CLIPScore stays stable across IDEFICS and IDEFICS (contextual).}
\label{table:subset_results_clipscore}
\end{table}

\section{Potential Use Cases of the CommVQA Dataset}

One potential use case for the CommVQA dataset is to assess whether a model performs better within a certain scenario than others. For instance, given a general-purpose VQA model, the CommVQA dataset provides a way to analyze whether this model excels at questions asked within the Shopping scenario, but struggles on questions asked within the Science Magazines scenario. These insights could prove useful for assessing when to deploy general-purpose VQA models versus specialized, domain-specific models.

Another use case is to evaluate a model’s ability to integrate contextual information along with the image. Prior work shows that people value visual explanations that incorporate contextual information \cite{beyondonesize, person_shoes_tree, muehlbradt2022s}. But how well do models integrate this contextual information in practice? In Figure \ref{fig:cosine_sim_plot}, we find that LlaVA and mPLUG-Owl both tend to repeat information from the description, while IDEFICS is more successful at integrating the contextual condition. CommVQA can help assess the ability of VQA models to integrate this contextual information, which as prior work has shown, is crucial for ensuring that model-generated outputs align with human preferences.

\section{Clarifying Context}

This section serves to clarify the role of context and how it is distinct from prior work. In our case, “context” means not only the scenario (i.e., the website where the image was encountered), but the scenario and the description. Prior work suggests that for blind and low vision users, this broader context should shape people’s informational needs about an image \cite{beyondonesize, person_shoes_tree}. Although other types of context have been shown to result in an improvement in the VQA and image captioning literature, this pragmatic framing of context is unique to our dataset. Our work bridges the gap between the studies showing that this version of context matters, and puts this observation into practice in generating a VQA dataset with a more realistic distribution of questions and answers.
\end{document}